\documentclass[a4paper,fleqn]{cas-dc}

\usepackage[numbers]{natbib}

\usepackage{svg}
\usepackage{float}
\usepackage{subfig}
\usepackage{amsmath}
\usepackage{amssymb}
\usepackage{amsthm}
\usepackage{xcolor}
\usepackage{graphicx}
\usepackage{geometry}
\usepackage{algpseudocode}
\usepackage{algorithm}

\definecolor{keyColor}{HTML}{0B5394}
\definecolor{todoColor}{HTML}{FF0000} 
\definecolor{outlineColor}{HTML}{7F6000}
\definecolor{outlineQuestionColor}{HTML}{FF6600}
\newcommand{\keysentence}[1]{\textcolor{black}{#1}}

\newcommand{\outlineA}[1]{}
\newcommand{\outlineQ}[1]{}

\def\tsc#1{\csdef{#1}{\textsc{\lowercase{#1}}\xspace}}
\tsc{WGM}
\tsc{QE}
\tsc{EP}
\tsc{PMS}
\tsc{BEC}
\tsc{DE}

\begin{document}
\let\WriteBookmarks\relax
\def\floatpagepagefraction{1}
\def\textpagefraction{.001}

\shorttitle{}

\shortauthors{Osburn et~al.}

\title [mode = title]{Systematic Constraint Formulation and Collision-Free Trajectory Planning Using Space-Time Graphs of Convex Sets}                      



%
\author[1]{Matthew~D.~Osburn}[
orcid=0000-0002-7316-858X
]



\ead{osburnm@byu.edu}



\author[1]{Cameron~K.~Peterson}[
orcid=0000-0002-6155-765X
]

\affiliation[1]{organization={Department of Electrical and Computer Engineering, Brigham Young University},
    city={Provo},
    state={Utah},
    country={United States of America}
    }

\author[2]{John~L.~Salmon}[%
   orcid=0000-0002-8073-3655
   ]


\affiliation[2]{organization={Department of Mechanical Engineering, Brigham Young University},
    city={Provo},
    state={Utah},
    country={United States of America}
    }






\begin{abstract}
In this paper, we create optimal, collision-free, time-dependent trajectories through cluttered dynamic environments. The many spatial and temporal constraints make finding an initial guess for a numerical solver difficult. Graphs of Convex Sets (GCS) and the recently developed Space-Time Graphs of Convex Sets (ST-GCS) enable us to generate minimum distance collision-free trajectories without providing an initial guess to the solver. We also explore the derivation of general GCS-compatible constraints and document an intuitive strategy for adapting general constraints to the framework. We show that ST-GCS produces equivalent trajectories to the standard GCS formulation when the environment is static, as well as globally optimal trajectories in cluttered dynamic environments. 
\end{abstract}


\begin{highlights}
\item Repeatable strategy for writing and adapting edge constraints for the GCS framework

\item First fully worked examples of deriving vertex and edge constraints for the GCS framework, something that is nonintuitive and not shown in the literature

\item Exploration of space-time graphs of convex sets (ST-GCS) and how space-time specific constraints allow the GCS framework to produce globally optimal minimum-distance trajectories in dynamic environments
\end{highlights}


\maketitle

\section{Introduction}

\begin{figure*}[!t]
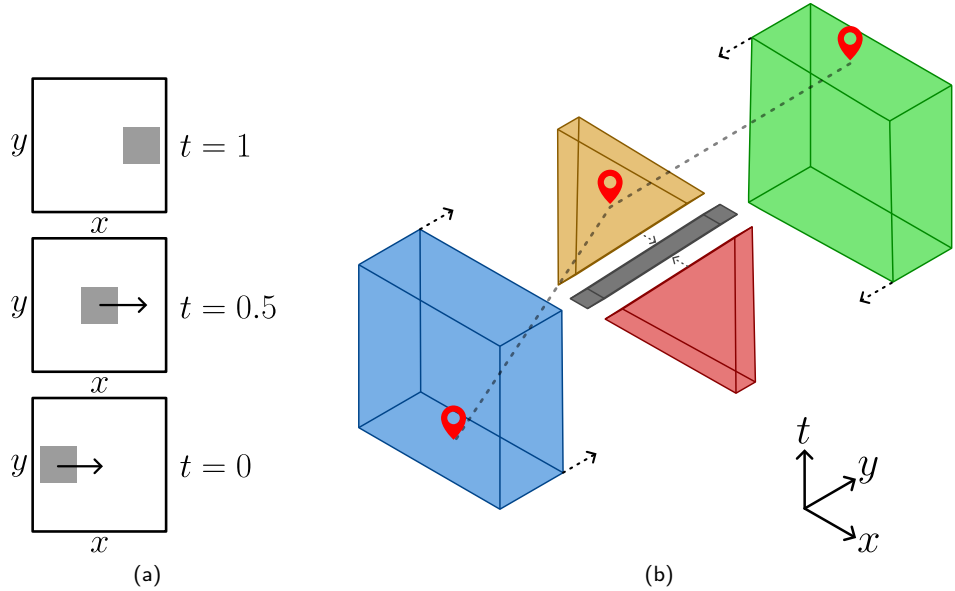

\centering
\subfloat[]{{\includesvg[width=0.2\linewidth]{figures/fig_time_crossection.svg} }}
\hspace{0.05\linewidth}
    \subfloat[]{{\includesvg[width=0.45\linewidth]{figures/fig_gcs_overview.svg} }}
    \caption{(a) A dynamic obstacle is shown at different times in $\mathbb{R}^2$. (b) Safe regions of space and time are partitioned into a Graph of Convex Sets (GCS), and then a collision-free trajectory is generated through the graph.} 
    \label{fig:gcs_intro_figure}
\end{figure*}

\keysentence{A fundamental problem in autonomous systems is generating a safe reference path or trajectory for a feedback controller to track.} Numerous planning methods exist for computing geometric paths that satisfy task-specific objectives~\cite{yang_path_2023}. However, such paths are typically insufficient in dynamic or time-varying environments~\cite{cai_mobile_2020,furze_robot_2022}, which require explicit modeling of how obstacles or goals change over time. In these cases, a trajectory—defined as a path parameterized over time—is more appropriate, as it enables temporal coordination and optimization in response to environmental changes~\cite{kroger_-line_2010,kacprzyk_path_2005}.

\keysentence{In motion planning, environments are frequently represented as discrete graphs by partitioning space into regions that enable efficient search-based methods.} Algorithms such as Dijkstra’s Algorithm~\cite{luo_surface_2020}, A*, and D*\cite{zhou_review_2022} are widely used to compute globally optimal paths without requiring an initial guess\cite{fink_globally_2019}. However, the computational complexity of graph search algorithms scales with graph size~\cite{martelli_complexity_1977,valenzano_worst-case_2014}, and the resulting paths often neglect physical constraints such as continuity or smoothness. As a result, discrete solutions typically require post-processing to convert them into feasible continuous trajectories~\cite{xiong_application_2021}, which may cause deviation from global optimality.

\keysentence{Continuous optimization methods, by contrast, can directly enforce physical constraints on trajectories.} These methods model obstacles using analytic functions such as signed distance fields~\cite{finean_predicted_2021}, control barrier functions~\cite{ames_control_2019, goncalves_safe_2024}, or polygonal representations~\cite{arai_avoidance_2024}, and they optimize a cost function subject to these constraints. However, a key limitation of continuous optimization is its reliance on a good initial guess. In complex or cluttered environments, poor initializations can lead to suboptimal or infeasible solutions. To mitigate this, sampling-based methods are often used to generate feasible initial trajectories~\cite{yuan_improved_2023,zhu_fast_2025,qureshi_intelligent_2015,tahir_potentially_2018}. Still, these approaches are especially sensitive to moving obstacles, which introduce additional temporal nonlinearities and local minima~\cite{osburn_optimization_2025,zang_unified_2024}.

To overcome these limitations, Marcucci \textit{et al.} introduced the Graphs of Convex Sets (GCS) framework~\cite{marcucci_graphs_2024,marcucci_shortest_2024,marcucci_motion_2022}. \keysentence{GCS is a trajectory optimization approach that integrates the discrete structure of graph search with the constraint handling capabilities of continuous optimization.} The framework constructs a graph of convex regions and formulates trajectory optimization as a combined graph search and convex optimization program. Notably, GCS has been applied to shortest-path planning~\cite{marcucci_shortest_2024,morozov_multi-query_2024}, minimum-time navigation in maze-like environments~\cite{marcucci_motion_2022}, and robotic arm motion planning in high-dimensional configuration spaces~\cite{cohn_non-euclidean_2024,garg_planning_2024,von_wrangel_using_2024}.

\keysentence{A key strength of GCS is that it finds continuous trajectories without requiring an initial guess.} It uses a branch-and-bound algorithm to iteratively refine candidate solutions until a globally optimal trajectory is found. Additionally, GCS provides metrics that indicate the quality of candidate solutions, allowing rapid selection of high-quality local optima~\cite{morozov_multi-query_2024}.

Despite these advantages, the current GCS pipeline has notable limitations. As explained in~\cite{marcucci_motion_2022}, it separates spatial and temporal planning by first computing a globally optimal collision-free path through a static environment, then determining the trajectory timing through an external nonlinear optimization. This latter step not only sacrifices global optimality guarantees but can also fail to produce a valid solution even when one exists. Moreover, existing approaches provide limited treatment of dynamic obstacle avoidance, which must be further developed and integrated before this framework can be reliably applied to safe trajectory planning. 

A particularly challenging aspect contributing to this complexity is the uniform formulation of constraints across the entire graph as formulated in ~\cite{marcucci_graphs_2024}. Even basic constraints, such as enforcing continuity between spline segments, must be expressed in a highly specific mathematical form, which is often abbreviated or omitted entirely. This requirement increases implementation challenges compared to other frameworks. To our knowledge, prior work has not provided explicit, accessible formulations of these constraints, further hindering adoption. This paper seeks to address this gap by presenting a detailed methodology for formulating and implementing such constraints, thereby facilitating broader use of the GCS framework.

One notable attempt at obstacle avoidance is by Werner et al.\cite{werner_superfast_2025}, who recompute the graph of convex sets and regenerate the trajectory at each timestep. To accelerate this process, their method requires GPU-accelerated graph computation. However, it does not incorporate any prior knowledge of obstacle motion into the planning, and still relies on a nonlinear optimization step to apply timing to the path—necessitating a good initial guess. In contrast, Marcucci et al.\cite{marcucci_motion_2022} propose joint planning for two robotic arms by concatenating their configuration spaces. While effective for a small number of agents, this approach scales poorly due to exponential growth in the combined configuration space, limiting its applicability. Additionally, because agents often cannot control the motion of other objects in real-world scenarios, such methods are unsuitable for general-purpose dynamic obstacle avoidance.

The most promising obstacle avoidance GCS formulation to date is that of Tang et al.~\cite{tang_space-time_2025}, who recently introduced the Space–Time Graph of Convex Sets (ST-GCS) framework for multi-robot motion planning. In this formulation, the configuration space is augmented with a time dimension, enabling constant-velocity obstacles to be represented in a manner compatible with the GCS framework. Their work considers only the other agents involved in joint planning between initial and final points in space. The method prioritizes minimum-time solutions and incorporates a mechanism to “reserve” space–time trajectories as dynamic obstacles for subsequent agents.

A limitation is that Tang et al. place conservative constraints on the final timing of the trajectory to ensure obstacle avoidance, potentially preventing the framework from finding true minimum time trajectories even if they exist. Additionally, their formulation is restricted to planning piecewise-constant velocity trajectories and is not applied to higher-order splines. Critical constraints that ensure causality within the space–time configuration space are present, but their importance is not discussed or elaborated.

In this paper, we generate collision-free, minimum-distance trajectories with a fixed end time using the ST-GCS framework, enabling direct comparison with the standard GCS formulation in static settings. The minimum distance cost function allows us to validate the framework’s applicability to dynamic scenarios and assess its performance and limitations within time-varying environments. Figure~\ref{fig:gcs_intro_figure} illustrates the ST-GCS formulation, depicting a moving two-dimensional object and a graph of convex sets generated in the time-augmented configuration space.

\keysentence{This paper makes two main contributions.}
First, we provide a systematic methodology, accompanied by step-by-step examples, for rewriting common constraints in the form required by the GCS framework.
Second, building on the ST-GCS framework, we investigate the generation of minimum-distance, collision-free trajectories in dynamic 2D environments. By assuming known obstacle positions and constant obstacle velocities, our approach produces time-dependent, higher-order optimal spline trajectories without relying on nonlinear solvers, initial temporal guesses, or incurring unnecessary computational complexity.

The remainder of this paper is structured as follows. Section~\ref{sec:background} provides an overview of the underlying mathematics for graphs of convex sets and Bézier Splines. Section~\ref{sec:methods} details our methodology and presents examples of formulating GCS-compatible constraints. Section~\ref{sec:results} presents the results of optimizing trajectories with GCS in dynamic environments.  Finally, Section~\ref{sec:conclusion} concludes the paper and discusses directions for future work.

\section{Background}\label{sec:background}

The GCS optimization framework offers robust guarantees for the end user and is grounded in concepts from convex analysis and optimization theory.  In this section, we will highlight the key concepts for understanding GCS at a high level.  For a more comprehensive understanding of GCS, we refer readers to~\cite{marcucci_graphs_2024,marcucci_shortest_2024,marcucci_motion_2022,morozov_multi-query_2024,cohn_non-euclidean_2024,garg_planning_2024,von_wrangel_using_2024,werner_superfast_2025,boyd_convex_2023}. 

\subsection{Convex Tools}

As the name suggests, convex sets are fundamental to the GCS framework. In this subsection, we introduce key definitions and equations regarding convex sets, but refer readers to chapters two and three of~\cite{boyd_convex_2023} for a more comprehensive treatment of the topic.   

A set  $\mathcal{S}~\subseteq~\mathbb{R}^n$ is convex if, for any two points within the set, the straight line segment connecting them lies entirely within $\mathcal{S}$.  Convexity can be demonstrated by taking two points $a,b \in \mathcal{S}$ and showing that the linear interpolation between the two always remains in the set $\mathcal{S}$. Formally, this is expressed as: 
\begin{equation}
    \begin{aligned}
        (1- \lambda)a + \lambda b \in \mathcal{S} \quad \forall a,b\in\mathcal{S} \quad \lambda \in [0, 1].
    \end{aligned}
    \label{eq:convex_set_def}
\end{equation}

\begin{figure}
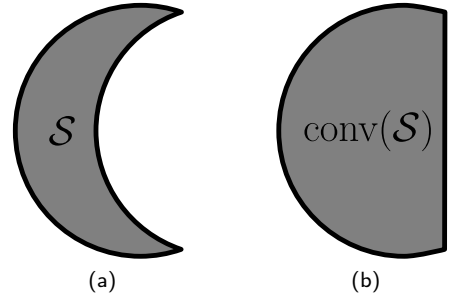

    \centering
    \subfloat[]{{\includesvg[width=2.25cm]{figures/fig_non_convex_set.svg} }}
    \hspace{1cm} 
    \subfloat[]{{\includesvg[width=2.25cm]{figures/fig_convex_hull.svg} }}
    \caption{(a) A nonconvex set $\mathcal{S}$. (b) The associated convex hull, $\text{conv}(\mathcal{S})$.}
    \label{fig:conv}
\end{figure}

Often, sets we encounter in practice are not convex.  The convex hull of set $\mathcal{S}$, denoted as conv($\mathcal{S}$),  is the smallest convex set that contains $\mathcal{S}$. An illustration of this is shown in Figure~\ref{fig:conv}.  If $\mathcal{S}$ is already convex, then conv($\mathcal{S}$)~=~$\mathcal{S}$.  

\begin{figure}
\centering\includesvg[width=3.5cm]{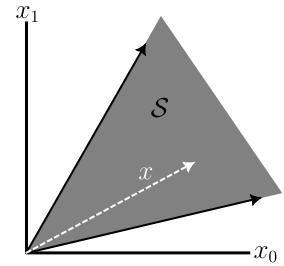}
    \caption{If $\mathcal{S}$ is a cone, the vector $x\in \mathcal{S}$ will still be in $\mathcal{S}$ after scaling by a non-negative constant.}
    \label{fig:cone}
\end{figure}

Another important concept in convex optimization is that of a cone.  A set $\mathcal{S}$ is called a cone (or said to be non-negative homogeneous) if for all $x \in \mathcal{S}$ and $\lambda \geq 0$ we have $\lambda x \in \mathcal{S}$ (shown in Figure~\ref{fig:cone}). A set is a convex cone if it is both convex and a cone. 

Notably, a set can be a cone without being convex.  For example, consider the union of the first and third quadrants of $\mathbb{R}^2$.  This set is a cone because any vector in the first and third quadrants of $\mathbb{R}^2$ can be scaled by a nonnegative number and remain within the set, but it is not convex.  
We will exclusively use convex cones in this paper.

Convex combinations are another useful tool we will use when translating constraints into the GCS framework. A convex combination $x_c$ of points $\vec{x}_1, \vec{x}_2,...\vec{x}_m \in \mathbb{R}^n$ is defined as 
\begin{equation}
    \begin{aligned}
        \vec{x}_c=\sum_{i=1}^{m} \vec{x}_i \lambda_i \quad \text{where} \quad\lambda_i \geq0, \quad\sum_{i=1}^{m} \lambda_i = 1.
    \end{aligned}
\end{equation} 
This guarantees that $\vec{x}_c$ lies within $\text{conv}\left(\{\vec{x}_1, \vec{x}_2,...,\vec{x}_m\}\right)$.  A linear combination of nonnegative coefficients can be normalized into a convex combination by dividing by the sum of the coefficients, assuming that the sum is positive.  Specifically, if 
\begin{equation}
    \begin{aligned}
        \lambda \vec{x}_c = \sum_{i=1}^{m} \vec{x}_i \lambda_i \quad \text{where} \quad\lambda_i \geq0,  \quad\lambda=\sum_{i=1}^{m} \lambda_i \label{eq:normal_convex_sum_1}
    \end{aligned}
\end{equation}
 then 
 \begin{equation}
    \begin{aligned}
        \vec{x}_c & = \frac{1}{\lambda}\sum_{i=1}^{m} \vec{x}_i\lambda_i \quad \text{where} \\  \lambda_i & \geq0, \quad \lambda > 0, \quad \frac{1}{\lambda}\sum_{i=1}^{m} \lambda_i = 1 . \label{eq:normal_convex_sum_2}
    \end{aligned}
\end{equation}
This relationship will be helpful when we derive and interpret the GCS-compatible continuity and differentiability constraints in the methods section.

\begin{figure}
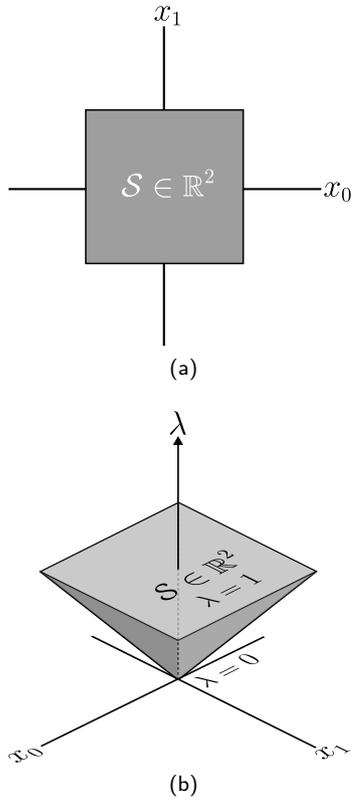

\centering
    \subfloat[]{{\includesvg[width=4.5cm]{figures/fig_set_in_r2.svg} }}
    \hfill
    \subfloat[]{{\includesvg[width=4.5cm]{figures/fig_set_in_r3_v2.svg} }}
    \caption{(a) A set $\mathcal{S}\in\mathbb{R}^2$. (b) the closure of the homogenization of $\mathcal{S}$. The $\text{clh}(\mathcal{S})$ forms a cone in $\mathbb{R}^3$.}
    \label{fig:clh}
\end{figure}

Another important tool used in the GCS framework is the closure of the homogenization of a closed set. Given a closed set $\mathcal{S}\in \mathbb{R}^{n}$, the closure of the homogenization of the set, denoted $\text{clh}(\mathcal{S})$, allows us to construct a cone in $\mathbb{R}^{n+1}$ (see Figure~\ref{fig:clh}). The $\text{clh}()$ operation is defined as 
\begin{equation}
    \begin{aligned}
        \text{ clh}(\mathcal{S}) :=& \left\{\text{concat}(\vec{x}, \lambda) \in \mathbb{R}^{n+1} : \lambda \geq0, \vec{x}\in \lambda\mathcal{S} \right\},
    \end{aligned}
\end{equation}
where $\text{concat}(\vec{x}, \vec{v})$ concatenates vectors $\vec{x} \in \mathbb{R}^n$ and $\vec{v} \in \mathbb{R}^m$ into $\mathbb{R}^{n+m}$. The original set can be recovered by taking a cross-section at $\lambda = 1$. If $\mathcal{S}$ is convex, then $\text{clh}(\mathcal{S})$ is also convex~\cite{marcucci_graphs_2024}.  In the GCS literature, constraints are represented using sets, generally written $\mathcal{X}$, and are applied across the entire graph using the $\text{clh}(\mathcal{X})$ representation.

\subsection{Graph Theory}

We denote a graph as $\mathcal{G} = (\mathcal{V}, \mathcal{E})$, where $\mathcal{V} = \{v_0, v_1, \dots, v_m\}$ is the set of vertices and $\mathcal{E} \subseteq \mathcal{V} \times \mathcal{V}$ is the set of edges. When $\mathcal{E}$ consists of ordered pairs, the graph is directed.  We will denote edges using ordered pairs $(v_a, v_b)$ to represent the directed connection from $v_a$ to $v_b$.  Some equations apply to every directed edge in the graph.  In this case, we will denote a generic edge in the graph as $e$, 

A graph of convex sets is a directed graph in which each vertex is associated with a convex set. Formally, it is written as $\mathcal{G}=(\mathcal{V}, \mathcal{E}, \mathcal{A})$, where $\mathcal{A}= \{A_{v_0}, A_{v_1}, ..., A_{v_m}\}$ is a collection of convex sets, with each set $A_{v_i}$ assigned to vertex $v_i\in \mathcal{V}$. These convex sets may reside in arbitrary dimensions and can encode a wide variety of problem structures.  Notably, the presence of an edge $(v_a, v_b) \in \mathcal{E}$ does not imply that the corresponding sets $A_{v_a}$ and $A_{v_b}$ intersect or overlap in $\mathbb{R}^n$.

\begin{figure}
    \centering
    \subfloat[]{{\includesvg[width=5cm]{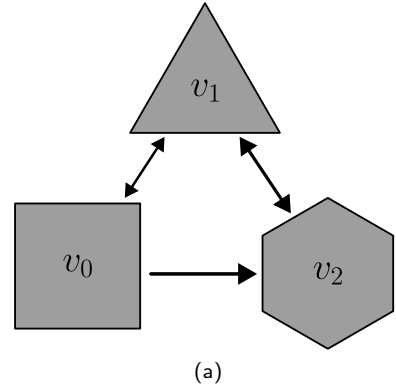} }}
    \caption{A directed graph of convex sets.}
    \label{fig:gcs_example_illustration}
\end{figure}

Figure~\ref {fig:gcs_example_illustration} shows an example graph of convex sets. The graph is directed, and each vertex is associated with a different convex shape in $\mathbb{R}^2$. In this illustration, the convex sets do not touch or overlap in $\mathbb{R}^2$.  

\subsection{Graphs of Convex Sets (GCS) Optimization}

The GCS optimization framework combines the strengths of continuous optimization and discrete graph searches. In this framework, a parameterized curve is assigned to each convex set. During optimization, curves corresponding to unvisited vertices are discarded, while curves at visited vertices connect at set boundaries to form a continuous trajectory. Unlike traditional graph searches, where the cost of an edge or vertex is a fixed value, the cost here varies continuously as the curves change within their respective convex sets.  Minimizing the cost of traveling through the graph can be formulated using the objective function:

\small
\begin{subequations}
\begin{align}
 \min_{x} \quad & \sum_{v \in \mathcal{V}} f_{v}(\vec{x}_{v}) + \sum_{e \in \mathcal{E}} f_{e}(\vec{x}_{a}, \vec{x}_{b}) \\
 \text{s.t.} \quad \notag \\ 
 & (\mathcal{V},\mathcal{E}, \mathcal{A}) \subseteq \mathcal{G} \\
&\vec{x}_{v} \in \mathcal{X}_{v} & \forall v\in \mathcal{V}\\
&\vec{x}_{a} \in \mathcal{X}_{e} & \forall e\in\mathcal{E}\\
&\vec{x}_{b} \in \mathcal{X}_{e} & \forall e\in\mathcal{E}.
\end{align}
\label{eq:gcs_optimization}
\end{subequations}\normalsize

Here, $x$ represents all the continuous variables defining the trajectory curve segments. The variables $\vec{x}_{v}$ are the design variables associated with the vertex $v$. The vertex cost function $f_{v}(\vec{x}_{v})$ represents the cost of the continuous variables within a vertex. The edge cost function $f_{e}(\vec{x}_{a}, \vec{x}_{b})$ represents the cost of traversing an edge from $v_a$ to $v_b$, and couples the continuous variables of the connecting vertices. The vertices $\mathcal{V}$, edges $\mathcal{E}$, and associated convex sets $\mathcal{A}$ that form the optimal trajectory are a subset of the graph $\mathcal{G}$ that minimizes the overall cost. The sets $\mathcal{X}_v$ and $\mathcal{X}_e$ define the vertex and edge constraints that the design variables must satisfy.  

A limitation of this formulation is that it does not provide a method to find $(\mathcal{V},\mathcal{E}, \mathcal{A}) \subseteq \mathcal{G}$ during optimization.  To address this, the problem can be reformulated as a mixed integer program (MIP) by introducing binary variables to represent the vertices and edges used in the final solution.  The binary variable $y_v \in \{0,1\}$ indicates whether a vertex is included in the solution, while the binary variable $y_e \in \{0,1\}$ indicates whether an edge is traversed.

The binary variables are multiplied into the cost and constraint terms of Equation~\ref{eq:gcs_optimization}.  This ensures that the cost and constraints are active only when the vertex or edge is traversed, and the constraint is removed when the vertex or edge is not in the final solution.  The MIP formulation is written as:

\small
\begin{subequations}
\begin{align}
\min_{x,y} \quad & \sum_{v \in \mathcal{V}} y_{v}f_{v}(\vec{x}_{v})  +  \sum_{e \in \mathcal{E}} y_{e}f_{e}(\vec{x}_{a}, \vec{x}_{b})\\
\textrm{s.t.} \quad \notag \\
& y_v \in \{0,1\}\\
& y_e \in \{0,1\}\\
& y_{v}\vec{x}_{v}\in y_{v}\mathcal{X}_{v} & \forall v\in \mathcal{V} \\
& y_{e}\vec{x}_{a} \in y_{e}\mathcal{X}_{e} & \forall e \in\mathcal{E} \\
& y_{e}\vec{x}_{b} \in y_{e}\mathcal{X}_{e} & \forall e \in\mathcal{E}. \label{eq:gcs_mip}
\end{align}
\end{subequations}\normalsize

As the graph grows, the combinatorial nature of the binary variables makes this MIP intractable to solve directly.  To address this, the binary variables are relaxed to continuous variables $y \in [0,1]$, and the problem is repeatedly solved using a branch-and-bound algorithm. In this formulation, the relaxed variables represent flow through the graph, from the start of the trajectory to its end.  The flow variables are constrained to ensure flow conservation at all intermediate vertices. Although the problem remains NP-hard, this relaxation is tight and significantly improves tractability. 

To handle the bilinear terms $y_{v}\vec{x}_{v}$ and $y_{e}\vec{x}_{a}, y_{e}\vec{x}_{b}$, a change of variable is applied: $\vec{z}_{v} = y_{v}\vec{x}_{v}$ and $\vec{z}_{e, v} = y_{e}\vec{x}_{v}$.  Assuming that $f_v$ and $f_e$ are convex functions and $\mathcal{X}_{v}$ and $\mathcal{X}_{e}$ are convex sets, this substitution transforms the problem into a mixed integer convex program (MICP), which drastically reduces the computation time per branch-and-bound iteration. The relaxed MICP can be written as:

\small
\begin{subequations} \label{eq:gcs_micp_}
\begin{align}
J(\vec{z},y) = &\sum_{v \in \mathcal{V}} y_{v}f_{v}\left(\frac{\vec{z}_{v}}{y_{v}}\right)  + \sum_{e\in \mathcal{E}} y_{e}f_{e}\left(\frac{\vec{z}_{a}}{y_{a}}, \frac{\vec{z}_{b}}{y_{b}}\right)\\
\min_{z,y} & \quad J(\vec{z},y) \\
\textrm{s.t.} \notag\\
  &   y_v \in [0,1]\\
 &y_e \in [0,1]\\ 
   & \sum_{} y_{(a,v)} + \delta_{s} = \sum_{} y_{(v,b)} + \delta_{f} = y_{v} \ \ \forall v \in \mathcal{V} \label{eq:gcs_micp:flow} \\
  & \vec{z}_{v}\in y_{v}\mathcal{X}_{v} \quad \forall v\in \mathcal{V} \label{eq:gcs_micp:clhv}\\
  & \vec{z}_{e, a} \in y_{e}\mathcal{X}_{e} \quad \forall e\in\mathcal{E} \label{eq:gcs_micp:clhea}\\ 
  & \vec{z}_{e, b} \in y_{e}\mathcal{X}_{e} \quad \forall e\in\mathcal{E}.\label{eq:gcs_micp:clheb}
\end{align}
\end{subequations}
\normalsize

Equation~\eqref{eq:gcs_micp:flow} introduces the flow conservation constraints. The variable $y_{(a,v)}$ denotes all the edge variables going to vertex $v$, and $y_{(v,b)}$ denotes all the edge variables leaving vertex $v$. The value $\delta_s = 1$ if $v$ is the starting vertex and $\delta_s = 0$ otherwise.  The value $\delta_f=1$ if $v$ is the final vertex and $\delta_f$ is $0$ otherwise.  These create a source and sink within the graph at the vertices that contain the start and end of the trajectory. 
In other papers, the variables $z$ and $y$ are concatenated in Equations~\eqref{eq:gcs_micp:clhv}, ~\eqref{eq:gcs_micp:clhea}, and~\eqref{eq:gcs_micp:clheb}, and the constraints are expressed using $\text{clh}(\mathcal{X})$. For clarity and consistency with the implementation in code, we omit this notation and leave the constraint as written here.  For a geometrical interpretation of the resulting set, we refer the reader to Figure~\ref{fig:clh}.

The previous GCS literature shows that the form of the MICP in Equation~\eqref{eq:gcs_micp_} can be used to find globally optimal trajectories. The convex relaxation is solved iteratively using a branch-and-bound algorithm, progressively adding constraints on the values of $y$ until the global optimal solution is found. We will leverage this formulation to find globally optimal collision-free trajectories.

\subsection{Generating The Graph Of Convex Sets}

There are several ways to generate convex sets. In this work, we use the Iterative Regional Inflation by Semidefinite programming (IRIS) algorithm, as described in~\cite{akin_computing_2015}.  This algorithm is the standard approach for generating convex sets in previous GCS literature.

The IRIS algorithm works by first sampling a point in the configuration space. Then the algorithm iteratively adds halfspace planes to define a convex set, as well as increases the volume of an ellipsoid inscribed within the proposed convex set, until the volume is maximized. If a point is sampled within an obstacle, it is rejected, and the convex set is not created.  

The IRIS algorithm can be configured to generate convex sets that either intersect or only touch at their boundaries. Once the free space is sufficiently sampled, each set is associated with a vertex in the graph, and the graph edges are created according to user-defined criteria. In this work, we generate sets that touch at their boundaries to avoid scenarios where the start and end points are contained within multiple sets.  Sets that touch at their boundaries are connected to form the edges of the graph.

This method is advantageous because it can generate collision-free convex sets in dimensions higher than $\mathbb{R}^3$, such as the configuration space of a 7-DOF robot arm. However, its main drawback is that it requires sampling the configuration space and does not guarantee complete coverage of the free space. Consequently, trajectories computed using this graph are globally optimal with respect to the graph but not necessarily globally optimal with respect to the entire free space.

\subsection{B\'ezier Splines}

Trajectories are often represented using piecewise polynomial functions known as splines.  Many types of splines exist, such as B-splines, minimum volume splines, and Bézier splines.  Previous GCS papers have used Bézier splines to represent an agent's trajectory, and for consistency, we also use Bézier splines.

Bézier splines are computationally efficient to sample and modify.  Their derivatives can be calculated easily as linear combinations of the control points defining the trajectory and are themselves lower-order splines.  Because the trajectory passes through the first and last control points of a Bézier curve, continuity and differentiability constraints are intuitive to enforce in many applications. 

The vector-valued equation for an $n^\text{th}$-order B\'ezier curve is 
\begin{align}
    \vec{B}(s) & = \sum_{k=0}^{n} \binom{n}{k}s^k(1-s)^{n-k}\vec{x}_k, & 0\leq s \leq 1,
\end{align}
where $\vec{x}_k$ is the $k$th vector-valued control point of the curve and $\binom{n}{k}$ is the binomial coefficient.  An $n^\text{th}$-order curve has $n+1$ control points. Individual Bézier curves are combined into a piecewise function to create a Bézier spline, which can be written as 
\begin{align}
    \vec{P}(r) & = \begin{cases} 
      \vec{B}_0(r) & 0\leq r < 1 \\
      \vec{B}_1(r-1) & 1\leq r< 2 \\
      \vdots \\
      \vec{B}_n(r-l) & l\leq r< l+1,  
   \end{cases}\label{eq:bezier_spline}
\end{align}
where $r\in[0,n]$ is the parameter that interpolates along the spline and $l$ denotes the number of individual Bézier curve segments.  

\begin{figure}
  \centering
  \includesvg[width=3.5cm]{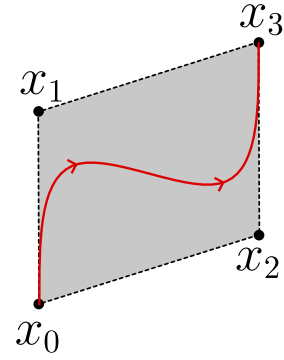}
    \caption{The spline (shown in red) stays within the convex hull of its control points when $s \in [0,1]$.}
    \label{fig_convex_hull_property}
\end{figure}

Bézier splines have a convex hull property~\cite{duncan_bezier_2005}, which ensures that a Bézier curve never leaves the convex hull of its control points (see Figure~\ref{fig_convex_hull_property}). In this illustration, the set of control points $C=\{x_0,x_1,x_2,x_3\}$ defines a Bézier curve which never leaves $\text{conv}(C)$.  This property is often used in collision avoidance algorithms to ensure safety.

In the GCS framework, a Bézier curve segment is assigned to each vertex of the graph.  The control points are constrained to remain within or on the boundary of their associated closed convex set, ensuring that the spline can never leave the set.  As the MICP problem is solved repeatedly, the curve segments eventually connect to form the final spline solution. Unused parts of the graph, where $y_v=y_e=0$, along with their associated Bézier curves, are discarded, leaving a single continuous spline traversing the graph of convex sets.

\begin{figure}
    \centering
    \subfloat[]{{\includesvg[width=4.5cm]{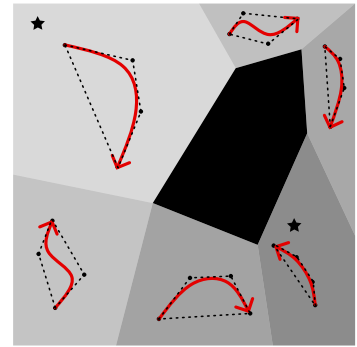} }\label{fig:gcs_example:initial}}
    \hfill
    \subfloat[]{{\includesvg[width=4.5cm]{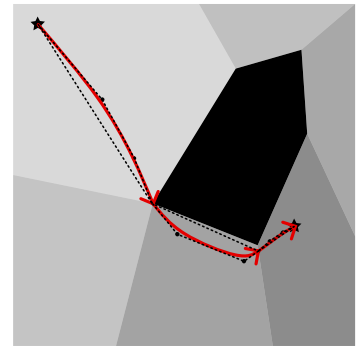} }\label{fig:gcs_example:final}}
    \caption{(a) B\'ezier curves associated with sets in GCS and (b) the solution to the GCS problem that connects B\'ezier curves to form a spline from start to end. Unused curves are removed.}
    \label{fig:gcs_example}
\end{figure}

Figure~\ref{fig:gcs_example:initial} illustrates an unsolved GCS trajectory problem.  The convex sets are depicted in various shades of grey, with the central obstacle shown in black.  Stars mark the start and end positions, the individual Bézier curves are drawn in red, and small black circles indicate the Bézier control points. Figure~\ref{fig:gcs_example:final} shows the solved trajectory, where the Bézier curve segments are connected to form a continuous path from the start to the goal, and unused curves are discarded.

\section{Methods}\label{sec:methods}

This section presents our methods for generating optimal collision-free trajectories using the GCS framework. We first describe how we generate convex sets to represent safe regions in the environment. We then outline a general strategy for formulating and implementing GCS-compatible constraints, addressing both vertex and edge constraints. We introduce the specific constraints used for trajectory planning in the space-time configuration space, including causality and kinematic constraints, as well as describe how obstacles are represented in the space-time configuration space. Finally, we conclude with the complete optimization formulation that enables time-dependent trajectories in 2D.

\subsection{GCS Constraints}

The formulation and implementation of constraints, particularly those spanning edges in the graph, are among the most challenging aspects of the GCS framework.  While prior work demonstrates that continuity and differentiability constraints can be implemented successfully within the GCS framework, the existing literature often lacks precise or accessible formulations of these constraints.  Particularly, prior work shows results that must have implemented constraints that fulfill 
Equations~\eqref{eq:gcs_micp:clhea} and ~\eqref{eq:gcs_micp:clheb}, but never show how this is done other than declaring the set $\mathcal{X}$ to represent the constraint.   We aim to address this gap by providing detailed derivations of these constraints as well as providing a repeatable step-by-step strategy to apply constraints across graph edges.

Trivial constraints, such as continuity and differentiability of the spline, become non-trivial within the convex relaxation of the mixed integer program (see Equation \eqref{eq:gcs_micp_}).  The convex relaxation raises questions such as: what does it mean for two Bézier splines in adjacent vertices to be continuous when the edge between them is only partially active (e.g., when $y_{e}=0.5$)?  What if multiple connecting edges are semi-active simultaneously?  How can a spline segment connect to multiple segments at once?  

Designing GCS-compatible edge constraints requires a different approach because constraints must adapt based on whether an edge is active, semi-active, or inactive. When an edge is active ($y_e=1$), the constraint must be fully enforced. For example, in spline continuity, this means the two curves in adjacent sets form a continuous piecewise function. When an edge is semi-active ($0 < y_e < 1$), the constraint must allow for a smooth transition between possible connections in the graph. Finally, when an edge is inactive ($y_e=0$), both sides of the constraint should become zero, effectively deactivating the constraint so it does not influence the design variables or their interactions across other edges. Designing constraints that satisfy these three behaviors while remaining continuous is a challenging task. 

Our method for writing GCS-compatible edge constraints is summarized in Algorithm~\ref{alg:edge_constraints}. The first step is to write the constraint as if it were active for a single edge in the solution subgraph.  Then, both sides of the equation are multiplied by the associated edge variable.  Next, the constraint is summed with respect to ingoing and outgoing edges and simplified using edge flow conservation. Then, the bilinear products for control points are replaced with the bilinear variable $z$.  Slack variables are introduced to represent the relationship across the edge and constrained to turn off when the edge turns off.   Finally, the constraint is repeated for each vertex in the graph. 

\begin{algorithm}
\caption{Writing Edge Constraints for GCS}
\begin{algorithmic}[1]
\State Write the constraint for a single edge, in terms of the control points within set $a$, $\vec{x}_a$, and the connecting control points within set $b$, $\vec{x}_b$.
\State Multiply both sides of the constraint by the edge decision variable $y_{(a,b)}$.
\State Sum the constraints for all outgoing edges connected to the vertex $a$ and separately sum the constraints for all ingoing edges connected to the vertex $b$.
\State Simplify the expression using the relationship in Equation~\eqref{eq:gcs_micp:flow}. Assume that $\delta_f=\delta_s=0$.
\State Substitute control point bilinear products, $y_v x_{v,i}$ with $z_{v,i}$ variables. Then introduce a substitution variable that represents the constraint relationship in step 2.
\State Simplify as needed.
\State Constrain the substitution variable as appropriate, ensuring that it turns off with respect to the edge decision variable $y_{(a,b)}$.
\State Repeat the constraint for each vertex $v\in \mathcal{V}$.
\end{algorithmic}
\label{alg:edge_constraints}
\end{algorithm}

The following subsections will give step-by-step derivations of the constraints that make up Equations~\ref{eq:gcs_micp:clhv},~\ref{eq:gcs_micp:clhea}, and ~\ref{eq:gcs_micp:clheb}.

\subsection{Convex Set Constraints}

Like many other GCS applications, we rely on the convex hull property to guarantee safety at every point along the entire trajectory. Recall that each vertex in the graph is associated with a convex set, within which a Bézier curve segment is defined. Because a Bézier curve always remains within the convex hull of its control points, we can guarantee that the curve stays within a safe region if we constrain the control points to lie within the associated convex set.  This constraint will be written in a way that is compatible with Equation~\ref{eq:gcs_micp:clhv}.

A polygonal convex set can be represented as the intersection of a finite collection of halfspaces.  Let $\hat{n}_i$ denote the normal vector of the $i^{th}$ plane and let $p_i$ be a point lying somewhere on that plane. A generic point $\vec{r}$ lies on or behind the halfspace plane if $\hat{n}_i \cdot \vec{r}\leq \hat{n}_i \cdot p_i$.  By stacking each face normal as a row vector in a matrix $A$, and stacking the corresponding dot products $\hat{n}_i \cdot  p_i$ into a vector $\vec{d}$, we can determine whether a point $\vec{r}$ lies within the convex set by checking the linear inequalities formed by $A\vec{r}\leq \vec{d}$. 

The variable $\vec{x}_{v,i}$ represents the $i^{th}$ control point in the curve associated with the vertex $v$.  The linear inequality $A_v\vec{x}_{v,i}\leq \vec{d}_v$ is applied to each control point in every set.  Rewriting this in terms of the bilinear variable $\vec{z}_{v,i} = y_v \vec{x}_{v,i}$ we get the equation
\begin{equation}
A_v \frac{\vec{z}_{v,i}}{y_v}\leq \vec{d}_v
\end{equation}
which can be rearranged to 
\begin{equation}
A_v \vec{z}_{v,i} \leq \vec{d}_v y_v.
\end{equation} 

\subsection{Continuity Constraints}

We begin by deriving the continuity constraints using Algorithm~\ref{alg:edge_constraints}. Our goal is for the last control point of a curve to match the first control point of the connecting curve when the edge is fully active.  Additionally, the constraint should allow for continuous transitions between possible connections when edges are only partially active. Because this constraint requires variables that are related by the edges in the graph, the following constraint formulation will fall under the category of Equations ~\ref{eq:gcs_micp:clhea} and ~\ref{eq:gcs_micp:clheb}.
 
We start by writing the constraint for a single generic edge in the graph:
\begin{equation}
\vec{x}_{a,n} = \vec{x}_{b,0}.
\end{equation}
Here, $\vec{x}_{a,n}$ denotes the last control point of the curve in set $a$, and $\vec{x}_{b,0}$ denotes the first control point of a connecting set $b$.

We then multiply both sides of the equation by the edge decision variable connecting set $a$ with set $b$, $y_{(a,b)}$, making the constraint fully active when the edge is included in the solution and inactive otherwise:
\begin{equation}
y_{(a,b)}\vec{x}_{a,n} = y_{(a,b)}\vec{x}_{b,0}. \label{cont_constraint_equality}
\end{equation}
Next, we sum these constraints over all outgoing edges connected to vertex $a$ and all ingoing edges connected to vertex $b$: 
\begin{align}\label{eq:edge_sums1}
\sum_{b}^{} y_{(a,b)}\vec{x}_{a,n} = \sum_{b}^{} y_{(a,b)}\vec{x}_{b,0}\\
\sum_{a}^{} y_{(a,b)}\vec{x}_{a,n} = \sum_{a}^{} y_{(a,b)}\vec{x}_{b,0}.\label{eq:edge_sums2}
\end{align}

We write the constraints in both directions to force the constraint slack variables to be applied properly on both sides of the edge. With only one side being accounted for, the constraint will only apply to the initial or final control point within a set.

Assuming the flow conservation constraint from Equation~\ref{eq:gcs_micp_}e holds so that
\begin{align}
y_{a} = \sum_{b}^{}y_{(a,b)} \label{flow_outgoing}\\
y_{b} = \sum_{a}^{}y_{(a,b)} \label{flow_ingoing},
\end{align}
the control points can be removed from the sums and Equation~\eqref{eq:edge_sums1} and Equation~\eqref{eq:edge_sums2} simplify to
\begin{align}
y_{a}\vec{x}_{a,n}  = \sum_{b}^{}y_{(a,b)}\vec{x}_{b,0} \label{unnormalized_outgoing_convex_hull} \\
y_{b}\vec{x}_{b,0}  = \sum_{a}^{}y_{(a,b)}\vec{x}_{a,n}.
\end{align}

\begin{figure}
    \centering
    \subfloat[]{{\includesvg[width=5cm]{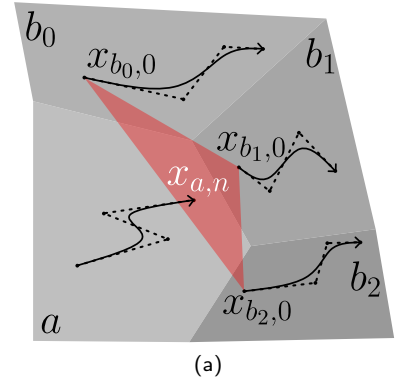} }}
    \hfill
    \subfloat[]{{\includesvg[width=5cm]{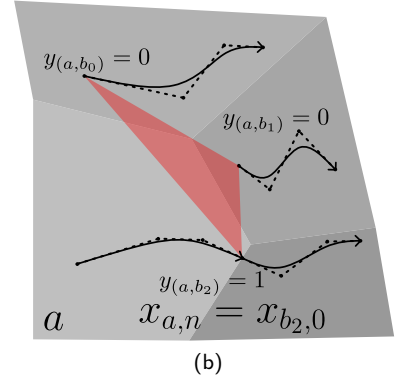} }}
    \caption{ (a) A vertex $a$ with adjacent vertices $b_i$. The continuity constraint ensures that the final Bézier control point falls in the convex hull of the connecting edges' first Bézier control point. (b) The two edges connect as $y_{(a,b_2)}\rightarrow1$. } 
    \label{fig:continuity_constraint}
\end{figure}

Referring back to Equations~\eqref{eq:normal_convex_sum_1} and~\eqref{eq:normal_convex_sum_2}, this constraint ensures that the last control point of a set lies within the convex hull of the first control points of the outgoing sets, and the first control point of a set lies within the convex hull of the last control points of the ingoing sets, weighted by their corresponding decision variables $y_{(a,b)}$.  This enforces continuity between curve segments when edges are active, enabling smooth transitions in the convex relaxation. Figure~\ref{fig:continuity_constraint} illustrates Equation~\eqref{unnormalized_outgoing_convex_hull}.


We then substitute the bilinear variables, $\vec{z}_{a,n} = y_{a} \vec{x}_{a,n}$, $\vec{z}_{b,0}=y_{b} \vec{x}_{b,0}$, to put the constraint in terms of the variables which represent the B\`ezier control points. From Equation~\eqref{cont_constraint_equality} we substitute  $\vec{p}_{(a,b)} = y_{(a,b)}\vec{x}_{a,n} = y_{(a,b)}\vec{x}_{b,0}$ to represent the point where both spline segments meet, and rewrite the constraint as:
\begin{align}
\vec{z}_{a,n}  &= \sum_{b}^{} \vec{p}_{(a,b)} \label{outgoing_cont_constraint_1}, \\
\vec{z}_{b,0}  &= \sum_{a}^{} \vec{p}_{(a,b)}.\label{ingoing_cont_constraint_1}
\end{align}

Finally, we constrain the point where the splines meet to lie on the intersection between the two convex sets, coupled by the edge decision variable $y_{(a,b)}$:


\begin{align}
\begin{bmatrix}
A_a \\ A_b
\end{bmatrix}
\vec{p}_{(a,b)}
\;\leq\;
\begin{bmatrix}
\vec{d}_a \\ \vec{d}_b
\end{bmatrix}
y_{(a,b)}.
\label{cont_constraint}
\end{align}


\subsection{Differentiability Constraints}

Next, we derive the differentiability constraint to ensure that the spline is differentiable with respect to the interpolation variable $r$ at the connection point between curves (see Equation~\eqref{eq:bezier_spline}). Similar to the continuity constraint, the differentiability constraint will fall under the category of Equations ~\ref{eq:gcs_micp:clhea} and ~\ref{eq:gcs_micp:clheb}.

For the path to be differentiable where the two curves connect, the vector between the last two control points of one curve must equal the vector between the first two control points of the connecting curve. For the edge $(a, b)$, this constraint can be written as: 
\begin{equation}
\vec{x}_{a,n} - \vec{x}_{a,n - 1} = \vec{x}_{b,1} - \vec{x}_{b,0}.
\end{equation}

As before, we multiply both sides of the equation by the edge decision variable $y_{(a,b)}$, making the constraint active when the edge is in the solution.  The equation can be written as:
\begin{equation}
y_{(a,b)}(\vec{x}_{a,n} - \vec{x}_{a,n - 1}) = y_{(a,b)}(\vec{x}_{b,1} - \vec{x}_{b,0}). \label{diff_constraint_equality}
\end{equation}

Next, we sum these constraints over all outgoing edges connected to vertex $a$ and all ingoing edges connected to vertex $b$:
\begin{align}
\sum_{b}^{} y_{(a,b)} (\vec{x}_{a,n} - \vec{x}_{a,n - 1}) = \sum_{b}^{} y_{(a,b)} (\vec{x}_{b,1} - \vec{x}_{b,0}) \\
\sum_{a}^{} y_{(a,b)} (\vec{x}_{a,n} - \vec{x}_{a,n - 1}) = \sum_{a}^{} y_{(a,b)} (\vec{x}_{b,1} - \vec{x}_{b,0}).
\end{align}

\noindent
Assuming the flow conservation constraint from Equation~\ref{eq:gcs_micp:flow} holds, the equations can be simplified and rearranged to

\begin{align}
 y_{a}(\vec{x}_{a,n} - \vec{x}_{a,n - 1}) &= \sum_{b}^{} y_{(a,b)} (\vec{x}_{b,1} - \vec{x}_{b,0}) \\
 y_{b}(\vec{x}_{b,1} - \vec{x}_{b,0}) &= \sum_{a}^{} y_{(a,b)} (\vec{x}_{a,n} - \vec{x}_{a,n-1}).
\end{align}

Then, we distribute the $y$ variables through the left-hand side 
and substitute the bilinear variables, $\vec{z}_{a,n} = y_{a} \vec{x}_{a,n}$, $\vec{z}_{a,n-1} = y_{a} \vec{x}_{a,n-1}$, $\vec{z}_{b,1} = y_{b} \vec{x}_{b,1}$, $\vec{z}_{b,0} = y_{b} \vec{x}_{b,0}$. After this, we use Equation~\eqref{diff_constraint_equality} and substitute $\vec{q}_{(a,b)} = y_{(a,b)}(\vec{x}_{a,n} - \vec{x}_{a,n - 1}) = y_{(a,b)}(\vec{x}_{b,1} - \vec{x}_{b,0}) $ to represent the difference between control points. The constraint can be written as:
\begin{align}
\vec{z}_{a,n - 1}  &= \vec{z}_{a,n} -\sum_{b}^{} \vec{q}_{(a,b)} \\
\vec{z}_{b,1}  &= \vec{z}_{b,0} +\sum_{a}^{} \vec{q}_{(a,b)}.
\end{align}

Finally, we constrain the 2-norm of $\vec{q}_{(a,b)}$ by some finite constant value $\gamma$, coupled with the edge decision variable,
\begin{align}
\left\lVert\vec{q}_{(a,b)}\right\rVert_2 \leq \gamma y_{(a,b)}.
\end{align}


\subsection{Space-Time Configuration Constraints}
We now introduce the space-time configuration constraints, which are essential for generating realistic trajectories in the space-time configuration space.  In the space-time configuration space, it is possible to construct splines that are non-causal and result in infinite velocities.  This happens when the time component of the trajectory is not unique, meaning that for certain values of $t$ there are multiple corresponding locations on the trajectory.  To prevent this, we impose a constraint that will enforce causality and limit the maximum velocity on the spline. This constraint will be written in a way that is compatible with Equation~\ref{eq:gcs_micp:clhv}

The maximum velocity of the spline $v_\text{spline}$ between control points can be bounded by the changes in the $x$, $y$, and $t$ components of the control points, so that
\begin{align}
v_\text{spline}\leq \frac{\lVert(\vec{x}_{v,i+1} - \vec{x}_{v,i})\rVert_{xy, 2}}{(\vec{x}_{v,i+1}-\vec{x}_{v,i})_t} &\leq v_{\text{max}} \\
 (\vec{x}_{v,i+1}-\vec{x}_{v,i})_t & \geq 0 
\end{align}
where $v_{\text{max}}$ is the maximum allowable velocity.  The notation $\lVert \cdot \rVert_{xy, 2}$ denotes the 2-norm with respect to just the $x$ and $y$ components of the control points, and $( \cdot )_{t}$ is the value of the time component within the parentheses.

\begin{figure}
    \centering
    \includesvg[width=3.5cm]{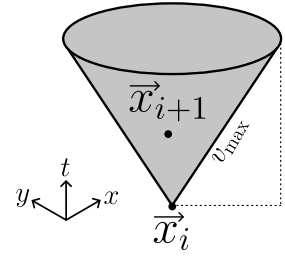}
    \caption{The velocity constraint forms a cone in $\mathbb{R}^3$. Each control point must lie within the cone formed by the previous control point to ensure the slope remains below the maximum velocity.}
    \label{fig:cone_constraint}
\end{figure}

\begin{figure}
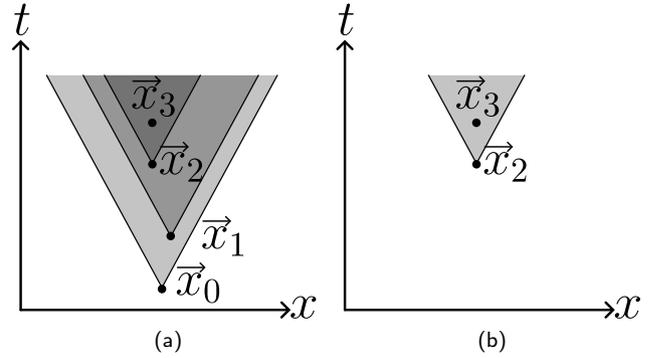

    \centering
    \subfloat[]{{\includesvg[width=4cm]{figures/fig_velocity_constraint_c0.svg} }}
    \hfill
    \subfloat[]{{\includesvg[width=4cm]{figures/fig_velocity_constraint_c1.svg} }}
    \caption{(a) The intersection of all velocity cone constraints remains convex. (b) The final control point must lie within this intersection, which is equivalent to just placing a velocity cone around the second-to-last control point.}
    \label{fig:cone_constraint_combined}
\end{figure}

\begin{figure*}
\centering
\subfloat[]{{\includesvg[width=0.20\linewidth]{figures/fig_time_crossection.svg} \label{fig:gcs_obstacle_construction:cross}}}
\hspace{0.1\linewidth}
    \subfloat[]{{\includegraphics[width=0.35\linewidth]{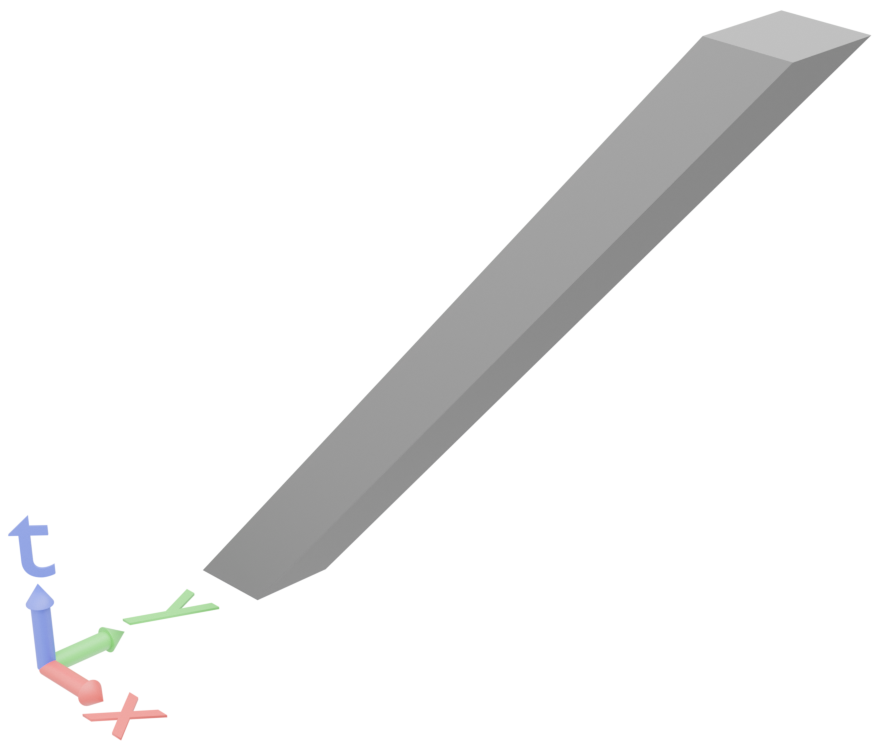} } \label{fig:gcs_obstacle_construction:start}}
    \caption{(a) A dynamic obstacle is shown at different times in $\mathbb{R}^2$. (b) A rendered view of the obstacle in $\mathbb{R}^3$.} 
\label{fig:gcs_obstacle_construction}
\end{figure*}

To make this constraint compatible with a convex solver, we rearrange it to be
\begin{align} 
\lVert(\vec{x}_{v,i+1} - \vec{x}_{v,i})\rVert_{xy, 2} & \leq v_{\text{max}}(\vec{x}_{v,i+1}-\vec{x}_{v,i})_t, \\
(\vec{x}_{v,i+1}-\vec{x}_{v,i})_t & \geq 0.
\end{align} 
Substituting the bilinear variable $\vec{z}_{v,i}$, this becomes
\begin{align}
\left\lVert \left(\frac{\vec{z}_{v,i+1}}{y_{v}} - \frac{\vec{z}_{v,i}}{y_{v}}\right)\right\rVert  _{xy, 2} & \leq v_{\text{max}}\left(\frac{\vec{z}_{v,i+1}}{y_v}-\frac{\vec{z}_{v,i}}{y_v}\right)_t \\
\left(\frac{\vec{z}_{v,i+1}}{y_v}-\frac{\vec{z}_{v,i}}{y_v}\right)_t  & \geq 0.
\end{align}
The constraint can be modified to 
\begin{align}
\left(\frac{\vec{z}_{v,i+1}}{y_v}-\frac{\vec{z}_{v,i}}{y_v}\right)_t \geq \epsilon,
\end{align}
which ensures that control points maintain a minimum time separation.

Finally, simplifying $y_v$ yields the constraint in its final form:
\begin{align}
\lVert(\vec{z}_{v,i+1} - \vec{z}_{v,i})\rVert_{xy, 2} & \leq v_{\text{max}}(\vec{z}_{v,i+1}-\vec{z}_{v,i})_t \\
(\vec{z}_{v,i+1}-\vec{z}_{v,i})_t & \geq y_v \epsilon.
\end{align}

The geometric intuition of this constraint is that it forms a cone in $\mathbb{R}^3$, depicted in Figure~\ref{fig:cone_constraint}a. Due to the convex hull property, the instantaneous slope of the Bézier curve never exceeds the slope made between two adjacent control points.  Therefore, if the slope between two control points is less than the maximum velocity, the curve will never violate the constraint.  To ensure that the maximum velocity constraint is satisfied, we must ensure that connecting adjacent control points is within the cone formed by the previous control point.  This is shown in Figure~\ref{fig:cone_constraint}.

While it is clear that this constraint is convex for any individual pair of adjacent control points, it may not be immediately obvious whether the combined constraints for all adjacent pairs remain convex. Figure~\ref{fig:cone_constraint_combined} illustrates that the feasible region for the final control point is the intersection of all the velocity cone constraints. This intersection of feasible regions is equivalent to just taking the velocity constraint at the previous control point. 

\subsection{Space-Time Obstacles}

Moving obstacles in $\mathbb{R}^2$ can be represented in the space-time configuration space by concatenating the initial positions of their vertices with the initial time, then concatenating the final positions with the final time, and finally connecting all vertices to form the obstacle in the space-time $\mathbb{R}^3$ configuration. Figure~\ref{fig:gcs_obstacle_construction} illustrates this process. This method assumes that obstacles are either static or move at constant velocity. More complex motions require finer sampling and will only approximate their true motion in $\mathbb{R}^3$.

\subsection{Cost}
The final component needed to implement the space-time GCS formulation is the cost function. It is important to choose the vertex cost function $f_v$ and the edge cost function $f_e$ so that the overall objective remains convex and simplifies when decision and relaxation variables are introduced, as shown in Equation~\ref{eq:gcs_micp_}.  In this work, we aim to minimize the distance traveled by the trajectory in the $(x,y)$ plane. A well-known property of B\`ezier curves is that the sum of the distances between adjacent control points is an upper bound on the curve's total length~\cite{gravesen_adaptive_1997}.  For simplicity, we assume that all convex sets touch in $\mathbb{R}^{n}$ and that there is no additional cost for traversing an edge, so that $f_e(x_e) = 0$.  Thus, the cost function for each node can be written in terms of the design variables as:
\begin{equation}
f_v(x_v) = \sum_{i=0}^{n-1} \lVert x_{v,i+1} - x_{v,i} \rVert_{xy, 2}.
\end{equation}
Substituting $z_{v,i}=y_{v}x_{v,i}$ into $f_v$ we get
\begin{equation}
\begin{aligned}
f_v\left(\frac{z_v}{y_v}\right) = & \sum_{i=0}^{n-1} \left\lVert \frac{z_{v,i+1}}{ y_v} - \frac{z_{v,i}}{{y_v}} \right\rVert_{xy, 2} \\ = & \frac{1}{\lvert y_v \rvert}\sum_{i=0}^{n-1} \rVert z_{v,i+1} - z_{v,i} \rVert_{xy, 2}.
\end{aligned}
\end{equation}
We can remove $y_v$ from the norm and, because it is non-negative, drop the absolute sign around it. The total cost function can be written as:
\begin{equation}
\begin{aligned}
J(y,z) =& \sum_{v \in \mathcal{V}} y_{v}\frac{1}{y_v}\sum_{i=0}^{n-1} \lVert z_{v,i+1} - z_{v,i} \rVert_{xy, 2}\\ J(z) = & \sum_{v \in \mathcal{V}}\sum_{i=0}^{n-1} \rVert z_{v,i+1} - z_{v,i} \rVert_{xy, 2}.
\end{aligned}
\end{equation}
This is the cost function that we used to generate collision-free trajectories using the GCS formulation.

\subsection{Optimization Formulation}

We incorporate the cost function along with the edge and vertex constraints into the overall optimization framework described by Equation~\ref{eq:gcs_micp_}. 
The resulting formulation is a relaxed mixed-integer convex program (MICP), written as:
\small
\begin{subequations}
\begin{align}
J(z) = & \sum_{v \in \mathcal{V}}\sum_{i=0}^{n-1} \rVert z_{v,i+1} - z_{v,i} \rVert_{xy, 2}\\
\min_{z,y,p,q} & \quad J(z)\\
\textrm{s.t.} \quad \notag\\
    & y_v \in [0,1]\\
& y_e \in [0,1]\\
    & \sum_{} y_{(a,v)} + \delta_{s} = \sum_{} y_{(v,b)} + \delta_{f} = y_{v}  \\
      & z_{s,0} = x_{initial} \\
  & z_{f, n} = x_{final} \\
  & A_v z_{v,i} \leq d_v y_v \\
  & \vec{z}_{a,n} = \sum_{b}^{} \vec{p}_{(a,b)}  \\
& \vec{z}_{b,0} = \sum_{a}^{} \vec{p}_{(a,b)} \\
  &\begin{bmatrix}
A_a \\ A_b
\end{bmatrix}
\vec{p}_{(a,b)}
\;\leq\;
\begin{bmatrix}
\vec{d}_a \\ \vec{d}_b
\end{bmatrix}
y_{(a,b)} \\
& \vec{z}_{a,n - 1}  = \vec{z}_{a,n} -\sum_{b}^{} \vec{q}_{(a,b)} \\
& \vec{z}_{b,1}  = \vec{z}_{b,0} +\sum_{b}^{} \vec{q}_{(a,b)} \\
& \left\lVert\vec{q}_{(a,b)}\right\rVert_2 \leq \gamma y_{(a,b)} \\
    & \lVert(z_{v,i+1} {-} z_{v,i})\rVert_{xy, 2} \leq v_{\text{max}}(z_{v,i+1}-z_{v,i})_t\\ 
  & (\vec{z}_{v,i+1}-\vec{z}_{v,i})_t \geq y_v \epsilon.
\end{align}
\label{eq:gcs_micp_final}
\end{subequations}
\normalsize

This final formulation includes relaxed binary variables $y$ for each vertex and edge, $n$ control point vectors $z_{v,i}$ within each vertex, and additional vectors representing continuity $\vec{p}_{(v,b)}$ and differentiability $\vec{q}_{(v,b)}$ for each directed edge. The variables $z_{s,0}$ and $z_{f,n}$ represent the first control point of the first set and the final control point of the final set, respectively. Each of these functions and constraints is convex and is compatible with general convex solvers.

\section{Results}\label{sec:results}

In this section, we apply our space-time GCS formulation to a range of scenarios and evaluate performance.  First, we will validate our formulation against the vanilla GCS formulation in a scenario with only static obstacles.  Then, we demonstrate its effectiveness in cluttered environments with containing many dynamic obstacles.

The simulations and algorithms were implemented in Python and executed on an AMD Ryzen 7 9700X processor with 32 GB of RAM. The convex optimization problems were formulated using CVXPY and solved with the open-source solver CLARABELL.  

\subsection{Standard GCS Comparison}

Because the standard GCS formulation is not compatible with dynamic environments, we will compare the space-time method against it in a static scenario.  We demonstrate that it produces the same optimal trajectory in this setting. The environment is decomposed into a graph of convex sets in two dimensions for the standard GCS formulation, and three dimensions for the space-time formulation. 

\begin{figure}
    \centering
    \subfloat[]{{\includesvg[width=5cm]{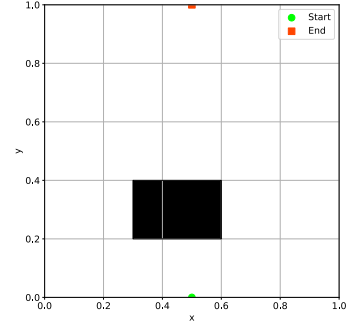} }}
    \hfill
    \subfloat[]{{\includesvg[width=5cm]{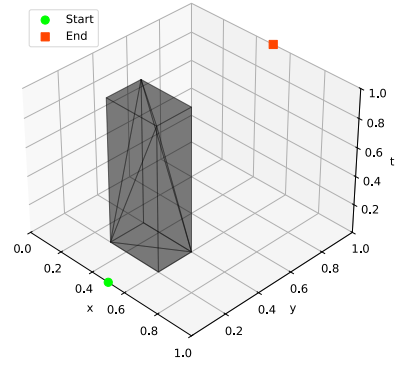} }}
    \caption{(a) The environment shown in 2D  and (b) an isometric view of the environment that will be used in the space-time GCS formulation.} 
    \label{fig:experiment_1_illustration}
\end{figure}

The scenario parameters are as follows.  An agent with state (x,y,t) begins at ($0.5m$, $0m$, $0s$) and must move to ($0.5m$, $1m$, $1s$).  The agent's maximum velocity is $2m/s$. A static obstacle is present with vertices at ($0.3m$, $0.2m$), ($0.6m$, $0.2m$), ($0.6m$, $0.4m$), and ($0.3m$, $0.4m$).  Because this obstacle is slightly off center, there is a single geometric minimum distance that avoids the obstacle between the start and end points, which is $ 1.0318 m$.  Figure~\ref{fig:experiment_1_illustration} depicts the scenario in 2d and 3d. The static obstacle is shown in black, with its vertices aligned vertically between $t=0s$ and $t=1s$. The start and end points are shown in green and red, respectively.  

\begin{figure}
    \centering
    \includesvg[width=5cm]{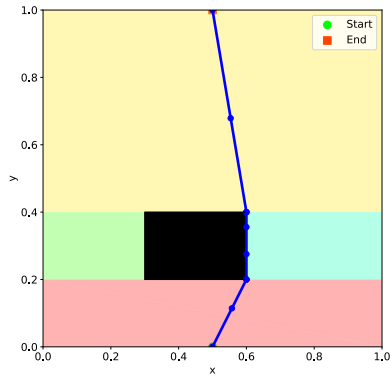}
    \caption{Path generated by the standard GCS formulation.} 
    \label{fig:experiment_1_2d}
\end{figure}

Figure~\ref{fig:experiment_1_2d} shows the path generated by the standard GCS formulation.  The brightly colored regions are the GCS convex sets. Control points are indicated by blue circles and are connected to adjacent control points by blue lines. The graph of convex sets consists of four sets and eight directed edges. The total computation time was $0.2s$. The optimal distance traveled is $1.0318m$, matching the geometric minimum.

\begin{figure}
    \centering
    \subfloat[]{{\includesvg[width=5cm]{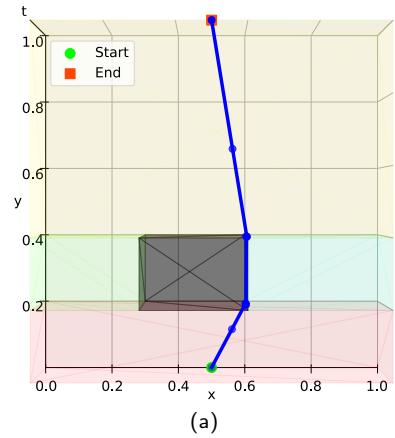}}}
    \hfill
    \subfloat[]{{\includesvg[width=5cm]{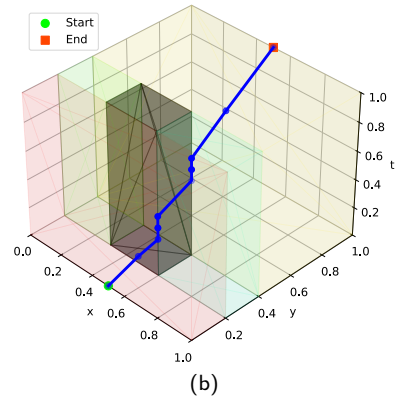}}}
    \caption{(a) Resulting sets and trajectory of the space-time GCS shown from the top and (b) an isometric view.} 
    \label{fig:experiment_1_3d}
\end{figure}

Figure~\ref{fig:experiment_1_3d} shows the connected B\'ezier control points generated by the space-time method from a top-down and an isometric view. Control points are indicated by blue circles and are connected to adjacent control points by blue lines.  When viewed from above, the geometric path is identical to the standard GCS solution.  The isometric view illustrates the vertical placement of each spline control point over time. The graph of convex sets consists of four sets and eight directed edges. The total computation time was $0.29s$. The optimal distance traveled in 2d is also $1.0318m$, which matches both the standard GCS solution and the true geometric minimum.  These results demonstrate that our modified GCS formulation produces trajectories identical to vanilla GCS in static environments, validating the correctness of our implementation.

\subsection{Dynamic Environment}

\begin{figure}
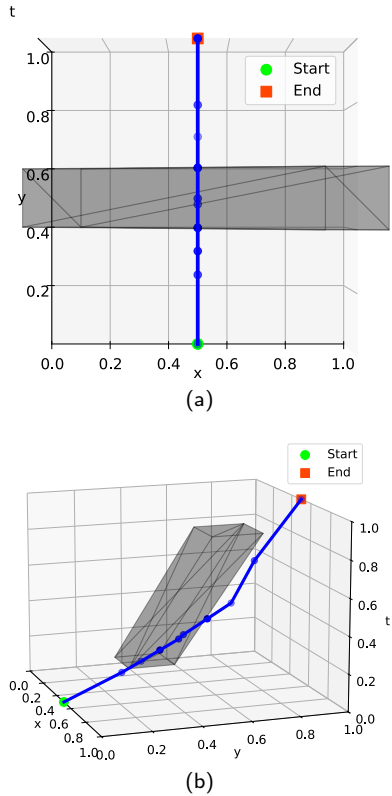

    \centering
    \subfloat[]{{\includesvg[width=5cm]{figures/fig_experiment_3_3d_top.svg}}}
    \hfill
    \subfloat[]{{\includesvg[width=5cm]{figures/fig_experiment_3_3d_iso.svg}}}
    \caption{(a) Resulting collision-free trajectory of the space-time GCS shown from the top and (b) in a perspective view.  The convex sets are omitted to visualize the trajectory clearly.} 
    \label{fig:experiment_3_3d}
\end{figure}

Next, we present a simple scenario with a single moving obstacle. In this case, multiple optimal trajectories have the same global minimum distance. We show that our method finds an optimal trajectory through the graph of convex sets while avoiding the moving obstacle. It should be noted that the standard GCS method can't be used to solve this scenario without introducing a non-linear solver to determine the timing of the path and avoid collisions.

The scenario parameters are as follows.  An agent with state $(x,y,t)$ starts at ($0.5m$, $0m$, $0s$) and must move to ($0.5m$, $1m$, $1s$).  The agent's maximum velocity is $2m/s$. There is a square-shaped dynamic obstacle with sides of length $0.2m$. The square starts centered at ($0m$, $0.5m$, $0s$) and moves to ($1m$, $0.5m$, $1s$).  The geometric minimum distance avoiding the obstacle from the agent's start location and end location is $ 1.0 m$.  

Figure~\ref{fig:experiment_3_3d} shows the connected B\'ezier control points generated by our method from a top-down and perspective view. Control points are indicated by blue circles and are connected to adjacent control points by blue lines.  When viewed from above, the geometric path moves in a straight line from start to finish.  The perspective view shows that the agent moves quickly in the beginning, as evidenced by the trajectory's slope, to move past the obstacle, then slows down. The graph of convex sets consists of four sets and eight directed edges. The total computation time was $0.52s$. The optimal distance traveled in 2D is also $1.0m$, which is the true geometric minimum distance. This example demonstrates the capability of our approach in generating collision-free, optimal trajectories in the presence of a dynamic obstacle, without relying on an initial guess.

\subsection{Monte-Carlo Simulations}

\begin{figure}
    \centering
    \includesvg[width=2.0in]{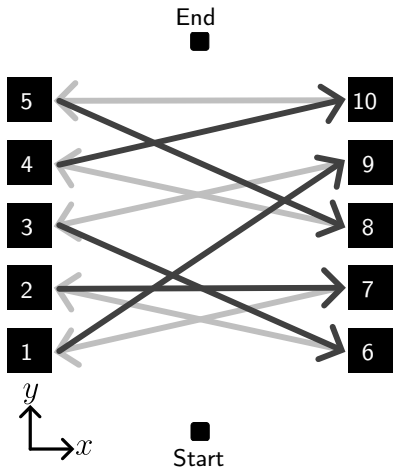}
    \caption{Illustration of the Monte-Carlo simulation scenario with 10 obstacles. The trajectory moves from the start point to the end while avoiding the obstacles moving across the middle.}
    \label{fig:montecarlo_scenario}
\end{figure}

\begin{figure}
    \centering
    \includegraphics[width=2.5in]{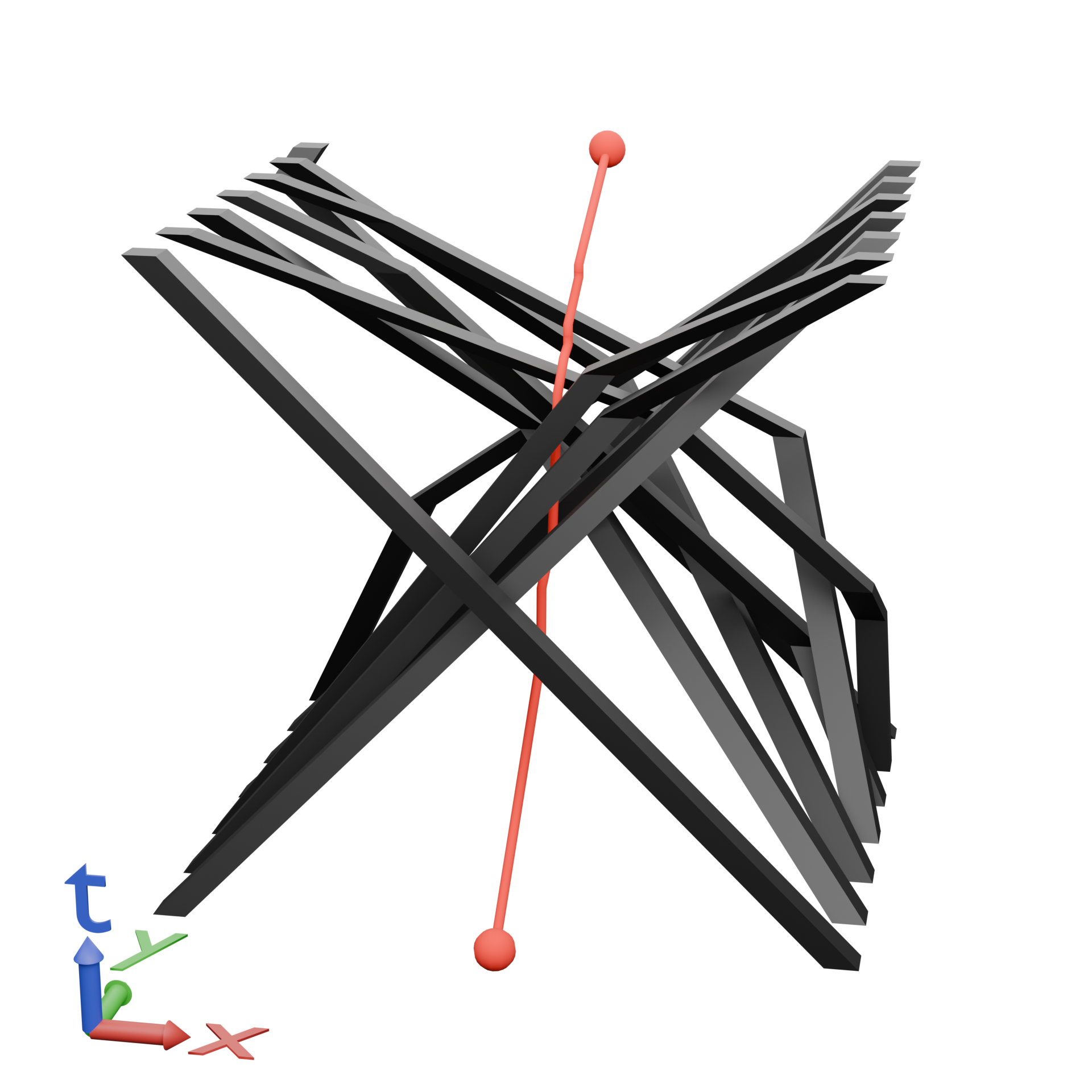}
    \caption{Rendered result of a minimum distance trajectory moving through a cluttered environment of 14 obstacles. Recall that the vertical direction represents time, meaning that this trajectory avoids spatial and temporal conflicts. This trajectory has a length of $1m$ in the $x-y$ plane.}
    \label{fig:montecarlo_traj}
\end{figure}

\begin{figure}
    \centering
    \includesvg[width=3in]{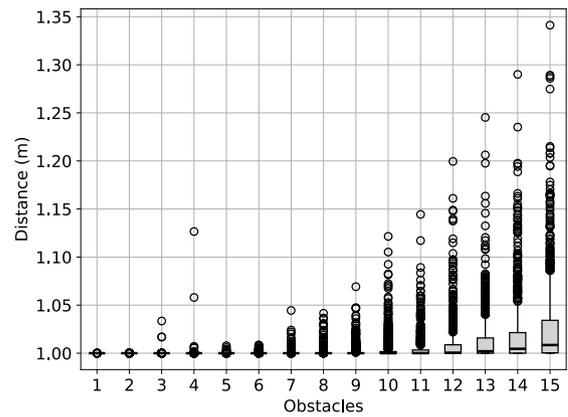}
    \caption{Monte-Carlo trajectory cost distributions.}
    \label{fig:montecarlo_cost}
\end{figure}

\begin{figure}
    \centering
    \includesvg[width=2.8in]{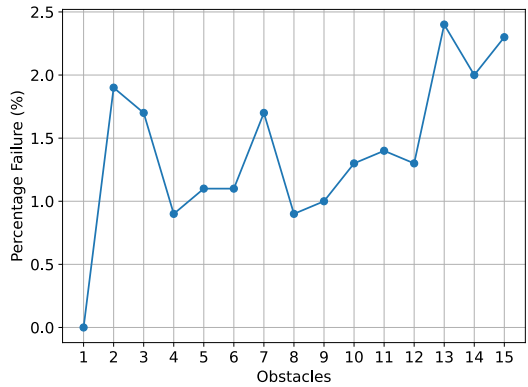}
    \caption{Percentage of failed trials as a function of obstacle count. Failures peak just below $2.5\%$ out of 1,000 trials. }
    \label{fig:montecarlo_failures}
\end{figure}

\begin{figure}
    \centering
    \includesvg[width=2.8in]{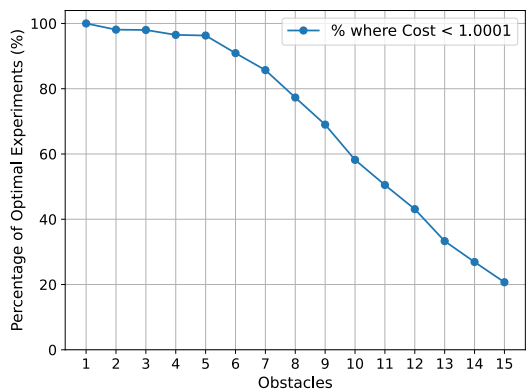}
    \caption{Count of Monte-Carlo runs achieving a cost within $0.01\%$ of the geometric minimum, as a function of obstacle count. The decrease reflects IRIS coverage limitations and obstacle interference.}
    \label{fig:montecarlo_optimal}
\end{figure}

To further evaluate the robustness of our space-time GCS formulation, we present a series of Monte-Carlo simulations in cluttered dynamic environments. An agent with state $(x,y,t)$ starts at ($0.5m$, $0m$, $0s$) and must move to ($0.5m$, $1m$, $1s$). The maximum allowable velocity is $3~\text{m/s}$, allowing the agent to traverse up to three times the straight-line distance within the available time.

Obstacles are incrementally introduced, ranging from one to fifteen in total. Each obstacle moves with a piecewise-constant velocity, beginning at evenly spaced positions along the lines $x=0$ and $x=1$, and ending at randomly assigned corresponding destinations on the opposite lines, $x=1$ and $x=0$, respectively.  Randomly selecting a corresponding final location forces many of the obstacles to cross in front of each other, resulting in a complicated collision-free space. Obstacles are arranged such that they never collide with each other during their motion. Figure~\ref{fig:montecarlo_scenario} depicts an illustration of the scenario with 10 obstacles. We varied the number of obstacles from 1 to 15 and conducted 1,000 randomized trials for each case. Randomization affects both the obstacle end-locations and the seed points used for IRIS-based convex set generation. To balance computation time and free-space coverage in the graph, we restrict the IRIS algorithm to generate at most $25$ convex sets per trial.

Figure~\ref{fig:montecarlo_traj} shows a trajectory generated using our formulation moving through a cluttered environment with 14 moving obstacles. Despite the density of obstacles, the agent finds a feasible path in the majority of trials, highlighting the flexibility of the ST-GCS framework. 

To quantify performance, we analyze the trajectory cost across all Monte-Carlo runs. Figure~\ref{fig:montecarlo_cost} presents a box-and-whisker plot of the cost distribution as a function of obstacle count. The horizontal bars represent the median value, the grey boxes represent the inter-quartile range, the vertical black lines represent results above the 25th percentile and below the 75th percentile, and the circles represent outliers. As expected, increasing the number of obstacles raises the median cost, reflecting the reduced availability of direct free-space corridors.  We consider trajectories shorter than $1.0001m$, within a percent of a percent of the best possible minimum, to be globally optimal.  The median value of each experiment is globally optimal up to and including $11$ obstacles.  

We also measure the failure rate of each experiment, defined as the percentage of runs where no feasible trajectory is returned. Failures occur due to a combination of: (1) insufficient connectivity of the graph formed by IRIS, (2) infeasibility arising from there not being a path through the dynamic obstacles, and (3) ill-conditioning in the convex solver. Figure~\ref{fig:montecarlo_failures} shows that for most scenarios the failure rate was between $1\%$ and $2\%$.  Because there is no trend in the error rate with respect to the number of obstacles, we infer that these failures were caused by ill-conditioning in the convex solver.  However, because we only measured failure for up to 15 obstacles, this does not mean the error rate may follow a trend as more obstacles are introduced. There were no experiments with a failure rate above $2.5\%$. 

Finally, we compare achieved trajectories against the theoretical geometric minimum path length. Figure~\ref{fig:montecarlo_optimal} shows the percentage of trials that we consider to be globally optimal (within $0.01\%$ of the $1m$). This percentage decreases as the number of obstacles increases, likely due to three factors: limited IRIS set coverage, obstruction of direct paths by obstacles, and the finite maximum velocity constraint.  

Overall, the Monte-Carlo simulations demonstrate that the proposed ST-GCS framework reliably generates collision-free trajectories in cluttered environments, with failure rates remaining low even under significant obstacle density. While trajectory optimality degrades with insufficient IRIS coverage and increasing clutter, these results confirm the framework’s ability to be applied in dynamic environments.

\section{Conclusion}\label{sec:conclusion}

We have presented a GCS formulation for trajectory generation in cluttered, dynamic environments. The constraint development strategy introduced here extends the applicability of GCS to a broader range of trajectory optimization problems. The space-time formulation enables GCS to handle cluttered dynamic environments effectively. Notably, this approach allows optimal trajectories to be generated without requiring a good initial spatial or temporal guess. 

Future work includes accounting for uncertainty in the positions of obstacles as well as adapting the space-time formulation to account for real vehicle dynamics. Overall, this work demonstrates a strategy to implement GCS constraints, as well as how the GCS framework can be extended to handle dynamic environments.

\section*{Acknowledgment}

This project was supported by NSF SBIR Phase 2 Award Number 2404858 and 4D Avionic Systems, LLC.  We thank Dr. Garth Thompson for his feedback and insightful suggestions.

\section*{Declaration of generative AI and AI-assisted technologies in the writing process.}

During the preparation of this work, the authors used ChatGPT to improve the readability and wording of different sections of the paper. After using ChatGPT, the authors extensively reviewed and edited the content to ensure accuracy and coherence.  We take full responsibility for the content of the published article.

\printcredits

\bibliographystyle{model1-num-names}

\bibliography{main}

\bio{figs/matt_picture_2}
Matthew Osburn received his B.S. degree in Electrical Engineering from Brigham Young University in 2023. He is currently pursuing a Ph.D. in Electrical Engineering at Brigham Young University. His research focuses on optimization, motion planning, and control theory, with a focus on their applications to unmanned aircraft systems.
\endbio

\bio{figs/CammyPeterson}
Cammy Peterson received her Ph.D. in Aerospace from the University of Maryland in 2012. She has over fifteen years of experience working as an engineer in the defense industry, most recently for Johns Hopkins University. She joined the Department of Electrical and Computer Engineering of Brigham Young University as an Assistant Professor in 2016. Her research interests include tracking, estimation, autonomy, and multi-vehicle motion coordination.
\endbio

\bio{figs/JohnSalmon}
John Salmon received his Ph.D. in Aerospace Engineering at the Georgia Institute of Technology.  His research interests include systems engineering, design, and integration, multi-disciplinary optimization, operations research, visual and data analytics, modeling and simulation, multi-agent multi-objective decision making, sports analytics, virtual reality, and uncertainty analysis.
\endbio

\end{document}